\documentclass[letterpaper]{article}
\usepackage[top=1in, bottom=1.25in, left=1.25in, right=1.25in]{geometry}

\usepackage{authblk}
\usepackage{amsmath}
\usepackage{amssymb}
\usepackage{amsfonts}
\usepackage{graphicx}
\usepackage{url}
\usepackage{natbib}

\usepackage{algorithm}
\usepackage{algorithmicx}
\usepackage{algpseudocode}
\usepackage{mathtools}
\usepackage{subfigure}
\usepackage{enumerate}
\usepackage{caption}
\usepackage{makecell}
\usepackage{cleveref}
\usepackage{multirow}

\begin{document}

\newcommand{\argmax}{\operatornamewithlimits{\arg \max}}
\newcommand{\argmin}{\operatornamewithlimits{\arg \min}}
\renewcommand{\algorithmicrequire}{\textbf{Input:}}
\renewcommand{\algorithmicensure}{\textbf{Output:}}
\newcommand{\btheta}{{\boldsymbol \theta}}
\newcommand{\bTheta}{{\boldsymbol \Theta}}
\newcommand{\bm}{{\mathbf{m}}}
\newcommand{\bh}{{\mathbf{h}}}
\newcommand{\calD}{{\mathcal{D}}}
\newcommand{\calJ}{{\mathcal{J}}}
\newcommand{\calM}{{\mathcal{M}}}
\newcommand{\defeq}{\stackrel{\mathrm{def}}{=}}

\title{Learning to Warm-Start \\Bayesian Hyperparameter Optimization}

\author[1]{Jungtaek Kim}
\author[2]{Saehoon Kim}
\author[1]{Seungjin Choi}
\affil[1]{\normalsize Pohang University of Science and Technology, Pohang, Republic of Korea}
\affil[2]{\normalsize AITRICS, Seoul, Republic of Korea}
\affil[ ]{\texttt{\{jtkim,seungjin\}@postech.ac.kr}, \texttt{shkim@aitrics.com}}

\maketitle

\begin{abstract}
Hyperparameter optimization aims to find the optimal hyperparameter configuration of a machine learning model, 
which provides the best performance on a validation dataset.
Manual search usually leads to get stuck in a local hyperparameter configuration, 
and heavily depends on human intuition and experience.
A simple alternative of manual search is random/grid search on a space of hyperparameters, 
which still undergoes extensive evaluations of validation errors in order to find its best configuration.
Bayesian optimization that is a global optimization method for black-box functions is now popular for hyperparameter optimization, 
since it greatly reduces the number of validation error evaluations required, compared to random/grid search.
Bayesian optimization generally finds the best hyperparameter configuration from random initialization without any prior knowledge.
This motivates us to let Bayesian optimization start from the configurations that were successful on similar datasets, 
which are able to remarkably minimize the number of evaluations.
In this paper, we propose deep metric learning to learn meta-features over datasets such that the similarity over them is effectively measured by Euclidean distance between their associated meta-features.
To this end, we introduce a Siamese network composed of deep feature and meta-feature extractors, 
where deep feature extractor provides a semantic representation of each instance in a dataset and meta-feature extractor aggregates a set of deep features to encode a single representation over a dataset.
Then, our learned meta-features are used to select a few datasets similar to the new dataset, 
so that hyperparameters in similar datasets are adopted as initializations to warm-start Bayesian hyperparameter optimization.
Empirical experiments on various image datasets (i.e., AwA2, Caltech-101, Caltech-256, CIFAR-10, CIFAR-100, CUB-200-2011, MNIST, and VOC2012) demonstrate our meta-features are useful in optimizing hyperparameters of convolutional neural networks for an image classification task.
\end{abstract}

\section{Introduction}
\label{sec:intro}

Machine learning model ${\calM}$ requires a choice of hyperparameter vector ${\btheta}$.
This hyperparameter vector ${\btheta}$ is chosen in the Cartesian product space of hyperparameter vectors $\bTheta = \Theta_1 \times \cdots \times \Theta_d$ where $\btheta = \left[\theta_1,\ldots,\theta_d\right]^{\top}$ and ${\theta_i \in \Theta_i}$ for ${1 \leq i \leq d}$. 
The vectors selected by machine learning experts using diverse experience and human intuition have guaranteed to find one of the best hyperparameter vectors which might show the best performance measure  (e.g., classification error and mean squared error).
However, such a na\"{i}ve hyperparameter search is not enough to find the hyperparameter vector that shows the best performance, even though we utilized the experience and intuition.
For this reason, many structured methods to find the better vector, such as random search, grid search~\citep{BergstraJ2012jmlr}, improvement-based search~\citep{KushnerHJ64,MockusJ78tgo}, and entropy search~\citep{hennig2012entropy} have been proposed.
In particular, various works~\citep{HutterF2011lion,BergstraJ2011nips,SnoekJ2012nips,Kim2016automl} which repeat the following steps: 
(i) predicting function estimates and their uncertainty estimates by a surrogate function model ${\calM_{\textrm{surrogate}}}$, 
(ii) evaluating domain space ${\bTheta}$ using outcomes of ${\calM_{\textrm{surrogate}}}$, 
and (iii) acquiring a new vector ${\btheta^{\dagger}}$, have been employed into hyperparameter optimization.

To measure which vector will be the best hyperparameter vector, the space needs to be either exploited or explored.
That trade-off of exploitation and exploration can be balanced by performance measure estimate and its uncertainty estimate, predicted by Bayesian regression models (e.g., Gaussian process regression~\citep{JonesDR98jgo} and Bayesian neural networks~\citep{springenberg2016bayesian}).
Such hyperparameter optimization that follows a framework of Bayesian optimization, referred to as \emph{Bayesian hyperparameter optimization} (BHO) finds the vector to maximize an acquisition function ${a(\cdot)}$:
\begin{equation*}
	\btheta^{\dagger} =  \argmax_{\btheta} a(\btheta | \mathcal{M}_{\textrm{surrogate}}).
	\label{eqn:max_a}
\end{equation*}
The details of BHO will be described in \Cref{subsec:bho}.

As summarized in \Cref{alg:bho}, ${k}$ initial vectors should be given in order to build a regression model.
Generally, random sampling methods such as na\"{i}ve uniform random sampling, Latin hypercube sampling~\citep{mckay1979comparison}, and quasi-Monte Carlo sampling are used to choose the initial vectors.
However, from exploitation and exploration trade-off perspective, initial vectors should be selected carefully, since better initial hyperparameter vectors encourage BHO to prevent exploration and focus on exploitation.
To find the better vectors, several works~\citep{BonillaEV2007nips,BardenetR2013icml,SwerskyK2013nips,YogatamaD2014aistats} have proposed the methods to share and transfer prior knowledge using block covariance matrices.
Moreover, hand-crafted and simply adjusted \emph{meta-features} over datasets, measured by dataset similarity~\citep{michie1994machine} have been proposed to transfer the prior knowledge to BHO~\citep{PfahringerB2000icml,FeurerM2015aaai,wistuba2015learning}.
The details of related works are described in \Cref{sec:related}.

\begin{algorithm}[t]
	\caption{Bayesian Hyperparameter Optimization}
	\label{alg:bho}
	\begin{algorithmic}[1]
		\Require Target function $\calJ(\cdot)$, $k$ initial hyperparameter vectors $\{\btheta_1^\dagger, \ldots, \btheta_k^\dagger\} \subset \bTheta$,
limit $T \in \mathbb{N} > k$
		\Ensure Best hyperparameter vector $\btheta^{\ast} \in \bTheta$
		\State Initialize an acquired set as an empty set
		\For {$i=1,2, \ldots, k$}
			\State Evaluate $\calJ_i \coloneqq \calJ(\btheta_i^\dagger)$
			\State Accumulate $(\btheta_i^\dagger, \calJ_i)$ into the acquired set
		\EndFor
		\For {$j=k+1,k+2,\ldots, T$}
			\State Estimate a surrogate function $\calM_{\textrm{surrogate}}$ with the acquired set $\{(\btheta_i^\dagger, \calJ_i)\}_{i=1}^{j-1}$
			\State Find $\btheta_j^\dagger = \argmax_{\btheta} a(\boldsymbol \theta | \mathcal{M}_{\textrm{surrogate}})$
			\State Evaluate $\calJ_j \coloneqq \calJ(\btheta_j^\dagger)$
			\State Accumulate $(\btheta_j^\dagger, \calJ_j)$ into the acquired set
		\EndFor
		\State $\textbf{return}$ $\btheta^{\ast} = \argmin_{\btheta_j^\dagger \in \{\btheta_1^\dagger,\ldots, \btheta_T^\dagger\}} \mathcal{J}_j$
	\end{algorithmic}
\end{algorithm}

In this paper, we propose the metric learning architecture to learn meta-features over datasets ${\bm}$, and apply the learned meta-features in BHO. 
We train deep feature extractor and meta-feature extractor over datasets via Siamese network~\citep{BromleyJ93nips}, 
matching meta-feature distance function ${d_{\textrm{mf}}(\cdot, \cdot)}$ with target distance function ${d_{\textrm{target}}(\cdot, \cdot)}$ (see \Cref{subsec:dist}).
After training the architecture, we determine $k$ datasets, comparing new test dataset with the datasets used in training the architecture.
Finally, the configurations of $k$-nearest datasets, which are previously measured, are used to initialize BHO.
To our best knowledge, this paper is the first work which proposes the method to warm-start BHO using the learned meta-features over datasets.

Before starting the main section, we summarize our contributions:
\begin{itemize}
	\item we visualize the effects of hyperparameters and datasets with respect to performance measure in order to reveal why we need to learn a meta-feature,
	\item we introduce a meta-feature extractor, trained by historical prior knowledge from the subsampled datasets obtained from various image datasets (i.e., AwA2, Caltech-101, Caltech-256, CIFAR-10, CIFAR-100, CUB-200-2011, MNIST, and VOC2012),
	\item we initialize BHO with ${k}$-nearest datasets, which the learned meta-features of new dataset decide on, and find the best hyperparameter vector following BHO steps.
\end{itemize}

\section{Background}
\label{sec:back}

In this section, we present Bayesian hyperparameter optimization and the reason why meta-features over datasets should be learned.
In addition, we describe some definitions, distance matching and target distance, used to learn meta-features.

\subsection{Bayesian Hyperparameter Optimization}
\label{subsec:bho}

Bayesian hyperparameter optimization (BHO), an application of Bayesian optimization~\citep{BrochuE2010arxiv} searches the best hyperparameter vector on the domain space ${\bTheta}$.
Suppose that we are given a dataset of training set and validation set $\calD = \{\calD_{\textrm{train}}, \calD_{\textrm{val}}\}$
with which we train a model involving hyperparameter vector ${\btheta}$.
Given a dataset $\calD$, the best hyperparameter vector is determined by minimizing
the validation error $\calJ(\btheta, \calD_{\textrm{train}}, \calD_{\textrm{val}})$.
As described in \Cref{alg:bho}, inputs to BHO are: 
(i) a target function  $\calJ(\btheta, \calD_{\textrm{train}}, \calD_{\textrm{val}})$
(whose functional form is not known in most of cases) which returns validation error or classification performance given 
hyperparameter vector and training/validation datasets, 
(ii) $k$ different hyperparameter vectors at initial design $\{\btheta_1^\dagger, \ldots, \btheta_k^\dagger\}$, 
and (iii) a limit $T$ which pre-specifies the number of candidates of hyperparameter vectors over which the best configuration is searched.
Then, the BHO undergoes the procedures which are explained below to return the best configuration of hyperparameters 
$\btheta^{\ast}$.

The BHO searches a minimum, gradually accumulating $(\btheta_j^\dagger, \calJ(\btheta_j^\dagger))$ with $j$ increasing.
Starting with a set of initial design $\{(\btheta_1^\dagger, \calJ_1), \ldots, (\btheta_k^\dagger, \calJ_k)\}$, 
a surrogate function model $\calM_{\textrm{surrogate}}$ is fit with the accumulated set of hyperparameter vector and its corresponding validation error.
In this paper, the Gaussian process (GP) regression model $\calM_{\textrm{GP}}$ serves as a surrogate function 
which approximates the landscape of $\calJ$ over the space $\bTheta$. 
The surrogate function well approximates the regions exploited so far, but has high uncertainty 
about the regions which are not yet explored.
Thus, rather than optimizing the surrogate function itself, the acquisition function $a(\btheta | \calM_{\textrm{GP}})$, which is constructed to
balance a trade-off between exploitation and exploration, is optimized 
to select the next hyperparameter vector at which the validation error $\calJ$ is evaluated. 
Assuming that the current GP has posterior mean $\mu(\btheta)$ and posterior variance $\sigma^2(\btheta)$, two popular acquisition functions
that we use in this paper are:
\begin{itemize}
	\item expected improvement (EI)~\citep{MockusJ78tgo}
	\begin{align*}
		a(\btheta | \mathcal{M}_{\textrm{GP}}) &= \mathbb{E}_{p\left(\mu\left(\btheta\right)|\btheta\right)}\left[ \max\{ 0, \mathcal{J}(\btheta^{\ddagger}) - \mu(\btheta) \} \right] \\
		&= ( \mathcal{J}(\btheta^{\ddagger}) - \mu(\btheta) ) \Phi \left( z(\btheta) \right) + \sigma(\btheta) \phi \left( z(\btheta) \right),
	\end{align*}
	where ${z(\btheta) = ( \mathcal{J}(\btheta^{\ddagger}) - \mu(\btheta) ) / \sigma(\btheta)}$, $\btheta^{\ddagger}$ is the best point known thus far, $\Phi(\cdot)$ denotes the cumulative distribution function of the standard normal distribution, and $\phi(\cdot)$ represents the probability density function of the standard normal distribution,
	\item GP upper confidence bound (GP-UCB)~\citep{SrinivasN2010icml}
	\begin{equation*}
		a(\btheta | \mathcal{M}_{\textrm{GP}}) = -\mu(\btheta) + \kappa \sigma(\btheta),
	\end{equation*}
	where $\kappa$ is a hyperparameter that balances exploitation and exploration to control the tightness of the confidence bounds.
\end{itemize}

\subsection{Distance Matching and Target Distance for Metric Learning}
\label{subsec:dist}

\begin{figure*}[t]
	\centering
	\subfigure[Caltech-256 (90\%)]{
		\includegraphics[width=0.47\textwidth, keepaspectratio]{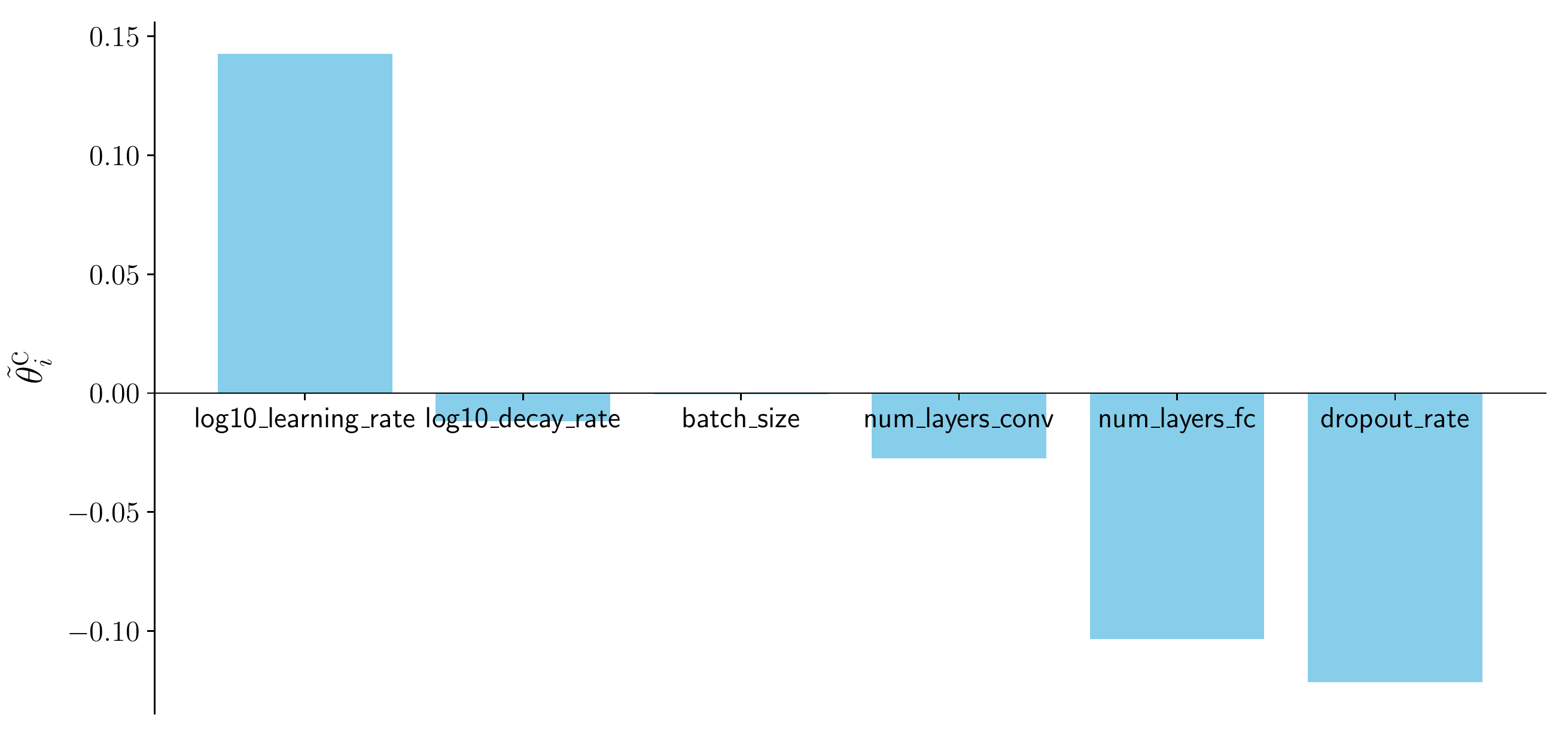}
		\label{fig:caltech256}
	}
	\subfigure[CUB-200-2011 (80\%)]{
		\includegraphics[width=0.47\textwidth, keepaspectratio]{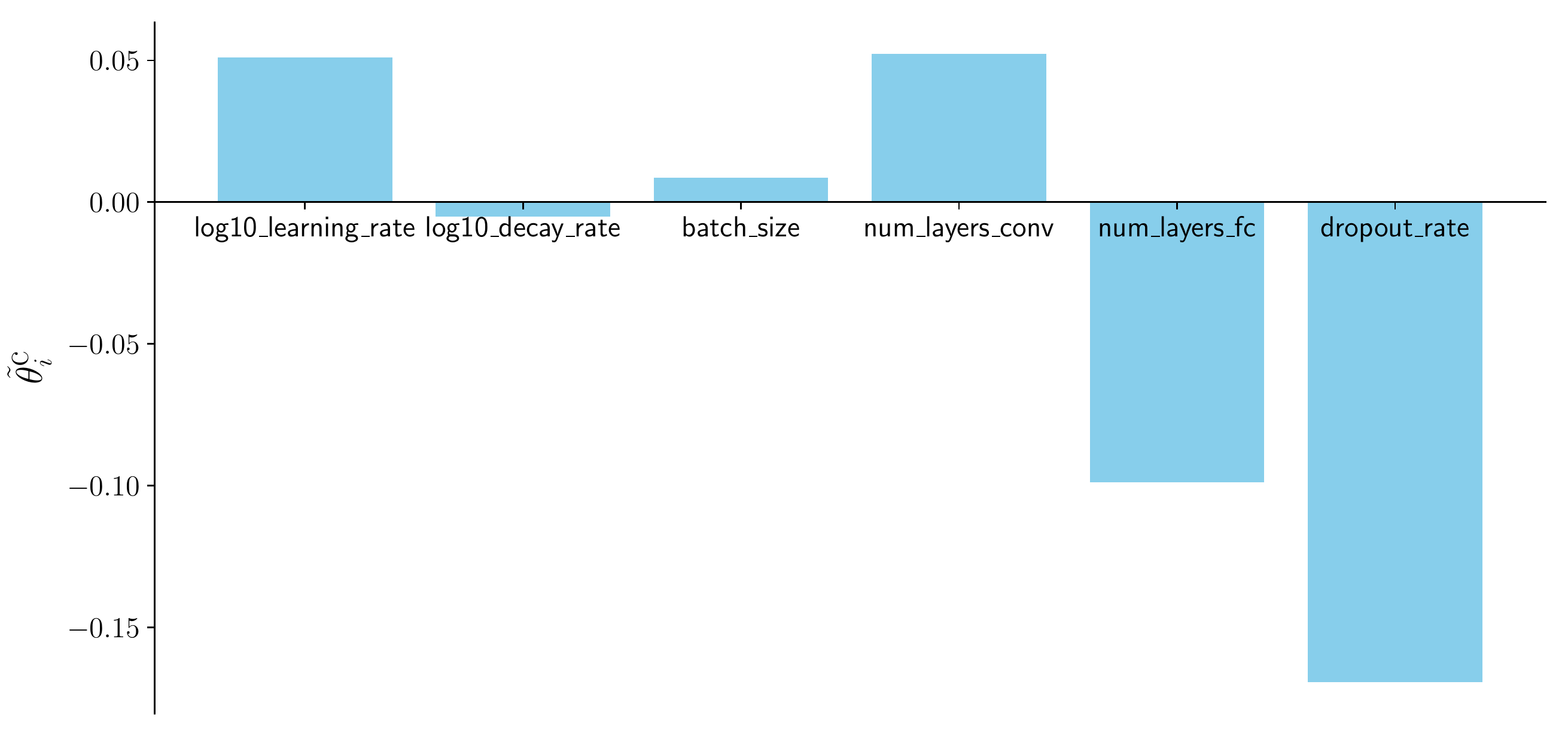}
		\label{fig:cub200}
	}
	\subfigure[MNIST (70\%)]{
		\includegraphics[width=0.47\textwidth, keepaspectratio]{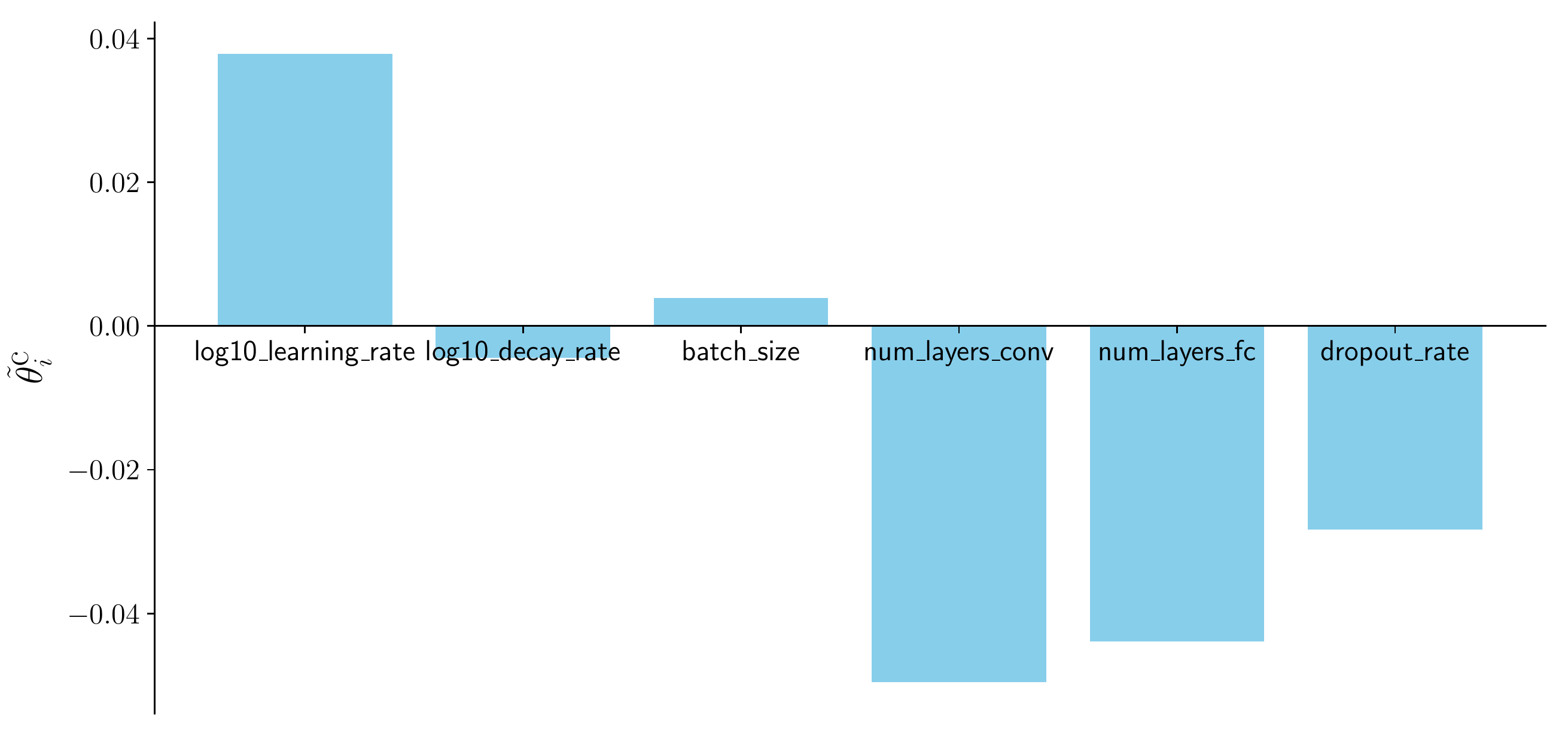}
		\label{fig:mnist}
	}
	\subfigure[VOC2012 (10\%)]{
		\includegraphics[width=0.47\textwidth, keepaspectratio]{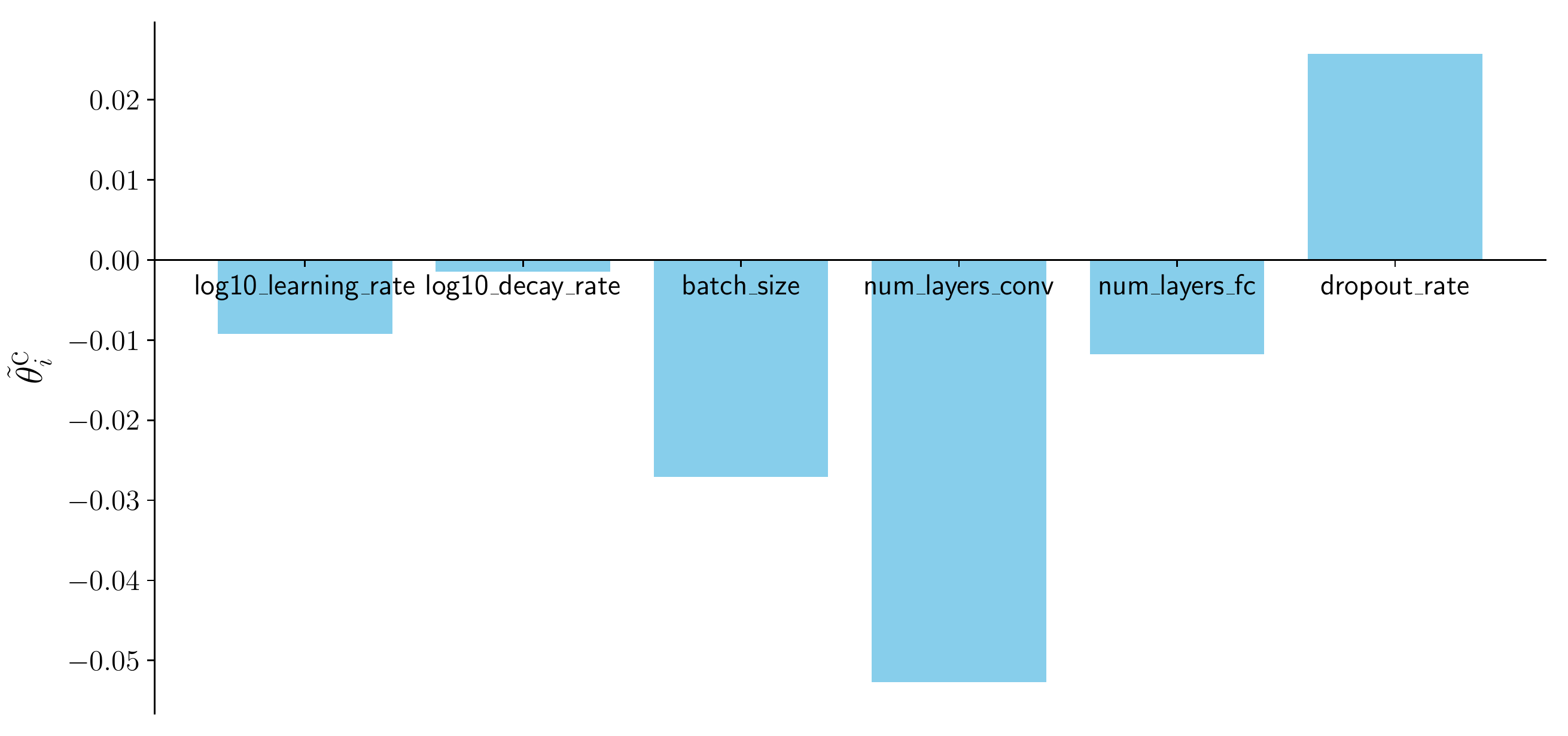}
		\label{fig:voc2012}
	}
	\caption{Subtracted CCoVs ${\tilde{\theta}_{i}^{c}}$ over hyperparameters for each dataset. ${i}$ in the subtracted CCoV indicates each hyperparameter dimension such as \textsf{log10\_learning\_rate}, \textsf{log10\_decay\_rate}, \textsf{batch\_size}, \textsf{num\_layers\_conv}, \textsf{num\_layers\_fc}, and \textsf{dropout\_rate}. \Crefrange{fig:caltech256}{fig:voc2012} show the subtracted CCoVs of each dataset. The details of datasets and hyperparameters are described in \Cref{subsec:setup_exp}.}
	\label{fig:dist}
\end{figure*}

Meta-feature vector ${\bm}$ produced by a meta-feature extractor ${\calM_{\textrm{mf}}}$ is used to measure meta-feature distance, 
which is matched to a target distance.
The model ${\calM_{\textrm{mf}}}$ is learned by minimizing a residual of target distance function ${d_{\textrm{target}}(\cdot, \cdot)}$ and meta-feature distance function ${d_{\textrm{mf}}(\cdot, \cdot)}$:

\begin{equation}
	\calM_{\textrm{mf}}^{\ast} = \argmin_{\calM_{\textrm{mf}}} \left\| d_{\textrm{target}}(\calD_i, \calD_j) - d_{\textrm{mf}}\left(\mathbf{m}(\calD_i|\calM_{\textrm{mf}}), \bm(\calD_j|\calM_{\textrm{mf}})\right) \right\|_2
	\label{eqn:residual}
\end{equation}
where ${\calD_i}$ and ${\calD_j}$ denote datasets compared, defined in \Cref{subsec:bho}.
${\calM_{\textrm{mf}}}$, which will be described in the subsequent section predicts the meta-feature vectors ${\bm_i \coloneqq \mathbf{m}(\calD_i|\calM_{\textrm{mf}})}$ and ${\bm_j \coloneqq \bm(\calD_j|\calM_{\textrm{mf}})}$, which indicate the outputs of ${\calM_{\textrm{mf}}}$ with respect to datasets ${\calD_i}$ and ${\calD_j}$, respectively.

Before introducing our deep meta-feature extractor ${\calM_{\textrm{mf}}}$, 
we first explain how target distance is measured and why meta-feature distance function is effective.
To measure target distance over datasets for metric learning, we need to collect prior knowledge from historical hyperparameter optimization.
Assuming that there are ${\{(\btheta_s, \calJ_s^{(t)})\}_{s=1}^{n}}$ for ${1 \leq t \leq K}$, where ${n}$ is the number of historical tuple of hyperparameter vector and validation error for each dataset, and ${K}$ is the number of the datasets which we have prior knowledge, 
we define a target distance function between two validation error vectors for the datasets ${\calD_i}$ and ${\calD_j}$, defined as ${L_1}$ distance:

\begin{align}
	d_{\textrm{target}}(\calD_i, \calD_j) &= \left\| \left[\calJ_1^{(i)} \cdots \calJ_n^{(i)}\right] - \left[\calJ_1^{(j)} \cdots \calJ_n^{(j)}\right] \right\|_1 \nonumber \\
	&= \sum_{s = 1}^n |\calJ_s^{(i)} - \calJ_s^{(j)}| \label{eqn:target}
\end{align}
where ${1 \leq i, j \leq K}$.

The target distance function, computed by historical validation errors, represents how different two datasets are.
For this implication, we can understand the target distance (computed by \Cref{eqn:target}) as the ground-truth pairwise distance.
However, there are two reasons why the target distance cannot be employed as the ground-truth distance directly:
(i) a distance for new dataset without prior knowledge cannot be measured, 
and (ii) a distance function, which can be interpreted as a mapping from hyperparameter space to validation error has a different multi-modal distribution with other mappings for different datasets.
The first reason is truly obvious, but we do not know yet whether or not a mapping from hyperparameter space to the error over datasets is multi-modal.

Since the dimension of hyperparameter space is higher than three-dimensional space, the mapping is difficult to visualize.
Thus, instead of visualizing directly, we adapt a concept of \emph{center of mass}.
We compute a coordinate of center of validation error (CCoV) ${\theta_{i}^{\textrm{c}}}$ for a dimension $i$:

\begin{equation}
	\theta^{\textrm{c}}_{i} = \frac{\sum_{s=1}^n {\tilde{\theta}}_{si} \calJ_s}{\sum_{s=1}^n {\calJ}_{s}}
	\label{eqn:ccov}
\end{equation}
where a normalized hyperparameter 
\begin{equation*}
	{\tilde{\theta}}_{si} = \frac{\theta_{si} - \min_{s=1, \dots, n} \theta_{si}}{\max_{s=1, \dots, n} \theta_{si} - \min_{s=1, \dots, n} \theta_{si}}
\end{equation*}
for ${1 \leq i \leq d}$ is provided.
Note that the superscript of ${\calJ_s^{(i)}}$ in \Cref{eqn:target} are dropped to simplify \Cref{eqn:ccov}.
Because \Cref{eqn:ccov} individually represents one dataset, it can be dropped.
As shown in \Cref{fig:dist}, each hyperparameter has different CCoV and those trends for each dataset are also different.
It implies that locations of modes are different over datasets.
To show the differences for all dimensions ${1 \leq i \leq d}$ clearly, we subtract ${0.5}$ from CCoV:

\begin{equation*}
	{\tilde{\theta}}_i^{\textrm{c}} = \theta_i^{\textrm{c}} - 0.5.
\end{equation*}

Given the subtracted CCoV ${\tilde{\theta}_i^{\textrm{c}}}$ such as \Cref{fig:dist}, 
the cases of which ${\tilde{\theta}_i^{\textrm{c}}}$ is larger than zero imply mode might be located in the region which has large absolute value (e.g., \textsf{log10\_learning\_rate} in \Crefrange{fig:caltech256}{fig:mnist}). 
If the original domain (e.g., \textsf{log10\_learning\_rate} and \textsf{log10\_decay\_rate}, see \Cref{tab:structure}) is negative, 
that means the location of mode is on the region which has small value.
On the other hand, if ${\tilde{\theta}_i^{\textrm{c}}}$ is smaller than zero, it implies mode might be located in the region that has small absolute value (e.g., \textsf{num\_layers\_conv} in \Cref{fig:mnist,fig:voc2012} and \textsf{num\_layers\_fc} in \Crefrange{fig:caltech256}{fig:voc2012}).
As shown in \Cref{fig:dist}, some trends of ${\tilde{\theta}_i^{\textrm{c}}}$ are different over datasets.
Furthermore, there are the cases which have different plus or negative sign (i.e., \textsf{log10\_learning\_rate}, \textsf{num\_layers\_conv}, and \textsf{dropout\_rate}).
Finally, based on these observations, we can argue the mappings from hyperparameter space to validation error have multi-modal distributions over datasets.

\section{Related Work}
\label{sec:related}

Hyperparameter optimization based on grid search or random search has been proposed~\citep{BergstraJ2012jmlr}.
Recently, various hyperparameter optimization methods based on Bayesian optimization or sequential model-based optimization have been proposed, 
and they demonstrate that Bayesian optimization or sequential model-based optimization performs better than grid search and random search with a small number of evaluations of validation error.
\citet{HutterF2011lion} suggest sequential model-based algorithm configuration method that models surrogate function as random forests.
\citet{SnoekJ2012nips} propose the integrated acquisition function, computed by Markov chain Monte Carlo estimation over acquisition functions varied by hyperparameters of GP regression.
\citet{BergstraJ2011nips} present Bayesian optimization which uses tree-structured Parzen estimator as a surrogate function modeling method.

Also, several works to meta-learn, transfer, or warm-start Bayesian optimization have been proposed.
A meta-learning~\citep{SchmidhuberJ87phd,ThrunS98book} method for Bayesian optimization is recently proposed~\citep{ChenY2017icml}.
Some works~\citep{BonillaEV2007nips,BardenetR2013icml,SwerskyK2013nips,YogatamaD2014aistats,poloczek2016wsc} capture and transfer the shared information between tasks using covariance functions in the context of GP regression.
In addition, the dataset similarity introduced in the literature~\citep{michie1994machine} is used in transferring a prior knowledge to BHO~\citep{PfahringerB2000icml,FeurerM2015aaai,wistuba2015learning}.
However, most of works, especially the methods based on the dataset similarity~\citep{PfahringerB2000icml,FeurerM2015aaai,wistuba2015learning} are not suitable to measure similarity between image datasets, because they measure the similarity with respect to hand-crafted statistics and simply learned landmark information.

From now on, we introduce the works related to our Siamese network with meta-feature extractors.
Various models with Siamese architecture, including CNNs~\citep{BromleyJ93nips}, MLPs~\citep{ChenK2011nips} and RNNs~\citep{MuellerJ2016aaai}, have been developed for deep metric learning since its first appearance in the work~\citep{BromleyJ93nips}.
All models have identical wings which extract meaningful features, and they are used to measure a distance between the inputs of wings.

Since in this paper we use two designs as a wing of Siamese network: (i) feature aggregation and (ii) bi-directional long short-term memory network (LSTM) (see \Cref{sec:model}), we attend to two structure of which the inputs are a set.
Feature aggregation method appeared in several works~\citep{edwards2016towards,charles2017pointnet,zaheer2017arxiv} takes a linear combination of feature vectors derived from instances in a set, and bi-directional LSTM for a set~\citep{vinyals2016matching} is also proposed.
They argue the tasks which take a set can be learned using such structures.
Moreover, \citet{zaheer2017arxiv} show that aggregation of feature vectors transformed by instances is invariant to the permutation of instances in a set.

\section{Proposed Model}
\label{sec:model}

\begin{figure*}[t]
	\centering
	\includegraphics[width=0.98\textwidth, keepaspectratio]{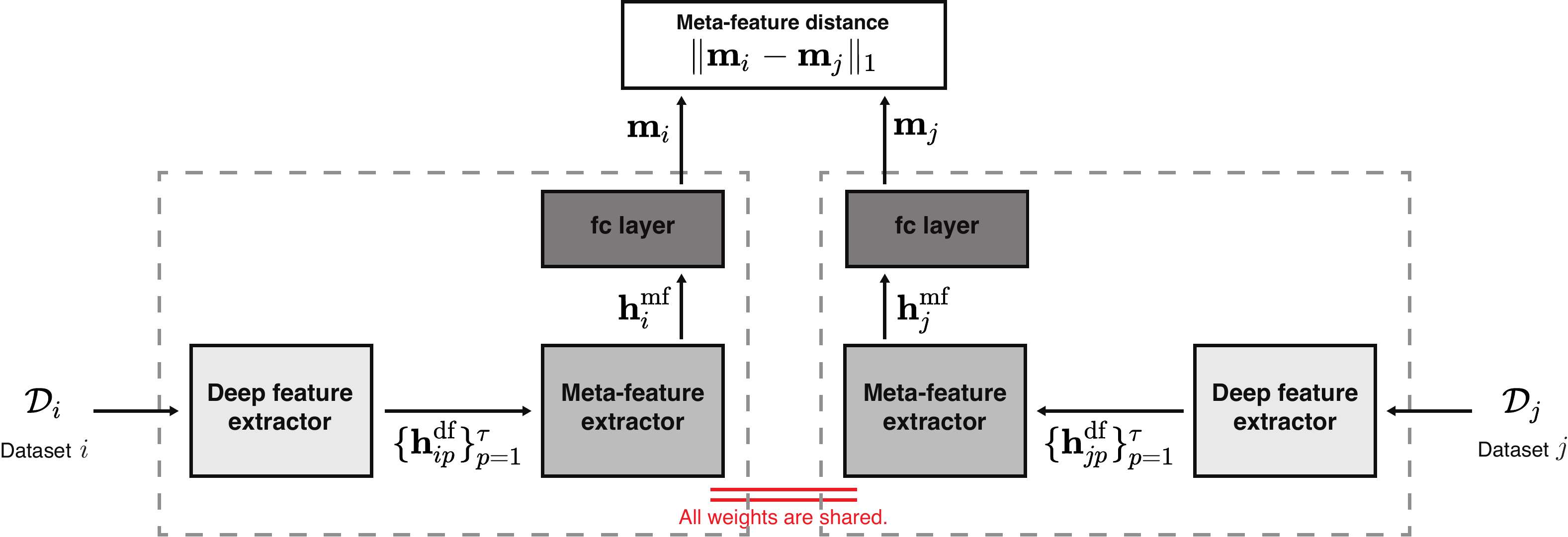}
	\caption{An illustration to show our Siamese network with deep feature extractor and meta-feature extractor. Deep feature extractor produces ${\{\mathbf{h}_{ip}^{\textrm{df}}\}_{p=1}^\tau}$ and ${\{\mathbf{h}_{jp}^{\textrm{df}}\}_{p=1}^\tau}$ for datasets ${\calD_i}$ and ${\calD_j}$, respectively. Next, meta-feature extractor produces ${\mathbf{h}_i^{\textrm{mf}}}$ and ${\mathbf{h}_j^{\textrm{mf}}}$, and the hindmost fully-connected layer (denoted as fc layer) produces ${\mathbf{m}_i}$ and ${\mathbf{m}_j}$. All shared weights are updated via minimizing the difference between meta-feature distance and target distance.}
	\label{fig:main}
\end{figure*}

We introduce our architecture composed of Siamese networks with deep feature and meta-feature extractors, as shown in \Cref{fig:main}.
Our network generates a deep feature of each instance in the dataset, and a set of deep features are fed into the meta-feature extractor to generate a meta-feature of dataset.

\subsection{Overall Structure of Siamese Network}
\label{subsec:siamese}

\begin{algorithm}[t]
	\caption{Learning a Siamese Network over Datasets}
	\label{alg:meta_features}
	\begin{algorithmic}[1]
		\Require A set of $n$ datasets ${\{\calD_1, \ldots, \calD_n\}}$, 
			target distance function ${d_{\textrm{target}}(\cdot, \cdot)}$, 
			number of subsamples in a dataset ${\tau \in \mathbb{N}}$, 
			number of iterations $T \in \mathbb{N}$
		\Ensure Deep feature extractor and meta-feature extractor (${\calM_{\textrm{df}}}$, ${\calM_{\textrm{mf}}}$) trained over ${\{\calD_1, \ldots, \calD_n\}}$
		\State Initialize ${\calM_{\textrm{df}}}$ and ${\calM_{\textrm{mf}}}$
		\For {$t = 1, 2, \ldots, T$}
			\State Sample a pair of datasets, i.e., $(\calD_i,\calD_j)$ for ${i \neq j}$, ${i, j = 1, \ldots, n}$
			\State Sample $\tau$ instances from each dataset in the pair $(\calD_i,\calD_j)$ selected above, to make ${|\calD_i| = |\calD_j| = \tau}$
			\State Update weights in $\calM_{\textrm{df}}$ and $\calM_{\textrm{mf}}$ using ${d_{\textrm{target}}(\calD_i,\calD_j)}$ via optimizing \Cref{eqn:residual}
		\EndFor
		\State \textbf{return} $(\calM_{\textrm{df}}, \calM_{\textrm{mf}})$
	\end{algorithmic}
\end{algorithm}

A Siamese network is used to learn a metric such that distance between meta-features of datasets is well matched to the target distance. 
Our Siamese network has two identical wings each of which is composed of deep feature and meta-feature extractors (denoted as $\calM_{\textrm{df}}$ and $\calM_{\textrm{mf}}$, respectively).
Specifically, the deep feature extractor transforms an instance of $\calD_i$ into $\bh^{\textrm{df}}_{ip}$ for $p=1, \ldots, n_i$, where $n_i$ is the number of instances in the dataset ${\calD_i}$.
The meta-feature extractor transforms a set of deep features $\{\bh^{\textrm{df}}_{ip} \}_{p=1}^{n_i}$ into a meta-feature of $\calD_i$, denoted as $\bm_i$.

Learning the proposed network requires the following inputs:
(i) a set of ${n}$ datasets ${\{\calD_1, \ldots, \calD_n\}}$, (ii) target distance function ${d_{\textrm{target}}(\cdot, \cdot)}$, (iii) the number of samples ${\tau}$ in a dataset, and (iv) the iteration number ${T}$.
\Cref{alg:meta_features} shows the procedure where deep feature and meta-feature extractors are iteratively trained by minimizing \Cref{eqn:residual}.
Specifically, we follow three steps in the iteration: 
(i) a pair of datasets is sampled from a collection of datasets (see \Cref{subsec:setup_exp}), 
(ii) ${\tau}$ instances are sampled from each dataset in the pair, 
and (iii) weights in $\calM_{\textrm{df}}$ and ${\calM_{\textrm{mf}}}$ are updated via optimizing \Cref{eqn:residual}.

Large $\tau$ (e.g., ${\tau \coloneqq n_i}$) makes learning procedure unstable and diverged.
However, simultaneously learning the networks with small $\tau$ seems to be converged to a local solution that poorly generalizes unseen instances.
Therefore, \Cref{alg:meta_features} should be expanded to the mini-batch version that updates weights via mini-batch gradient descent. 
The mini-batch version collects a set of ${\tau}$ instances by repeating Line 4 of \Cref{alg:meta_features}, and updates weights via mini-batch gradient descent.

\subsection{Meta-Feature Extractor}
\label{subsec:wing}

Before explaining the details of meta-feature extractors, we first introduce the issues about handling instances of datasets.
Dataset is a set of instances (i.e., images in this paper) which is invariant to the order of instances and varies the number of instances for each dataset.
Therefore, it is difficult to feed in deep neural networks as well as common shallow classifiers, which have fixed input size.
To resolve these problems, we consider two designs into meta-feature extractor:
\begin{itemize}
	\item aggregation of deep features (ADF): deep features ${\bh^{\textrm{df}}}$, derived from deep neural networks (e.g., convolutional layers and fully-connected layers) are aggregated as summation or arithmetic mean of them:
	\begin{equation*}
		\bh^{\textrm{mf}} \coloneqq \bh_{\textrm{ADF}} = \sum_{p=1}^{\tau} \bh^{\textrm{df}}_p
		\quad \textrm{or} \quad
		\frac{1}{\tau} \sum_{p=1}^{\tau} \bh^{\textrm{df}}_p,
		\label{eqn:adf}
	\end{equation*}
	\item bi-directional long short-term memory network (Bi-LSTM): the deep features ${\bh^{\textrm{df}}}$ are fed into Bi-LSTM. Bi-LSTM can be written as
	\begin{equation*}
		\bh^{\textrm{mf}} \coloneqq \bh_{\textrm{Bi-LSTM}} = \textrm{Bi-LSTM}(\bh^{\textrm{df}}_{1:\tau}),
		\label{eqn:bi_lstm}
	\end{equation*}
	where ${\bh^{\textrm{df}}_{1:\tau}}$ denotes ${[\bh^{\textrm{df}}_{1}, \bh^{\textrm{df}}_{2}, \ldots, \bh^{\textrm{df}}_{\tau-1}, \bh^{\textrm{df}}_{\tau}]}$.
\end{itemize}

As we mentioned in \Cref{sec:related}, they do not depend on the number of instances, and also they tend not to be affected by the order of instances. 
In order to learn a meaningful weights of meta-feature extractor, we input deep features ${\bh^{\textrm{df}}}$ into meta-feature extractor (e.g., ADF or Bi-LSTM).
The deep features obtained after passing convolutional and fully-connected layers with non-linear activation function can be used.
Finally, fully-connected layer, followed by the output of meta-feature extractor produces a meta-feature vector, as shown in \Cref{fig:main}.
The details and practical configuration of meta-feature extractors will be described in the subsequent section.

\section{Experiments}
\label{sec:exp}

\begin{figure*}[t]
	\centering
	\subfigure[ADF]{
		\includegraphics[width=0.30\textwidth, keepaspectratio]{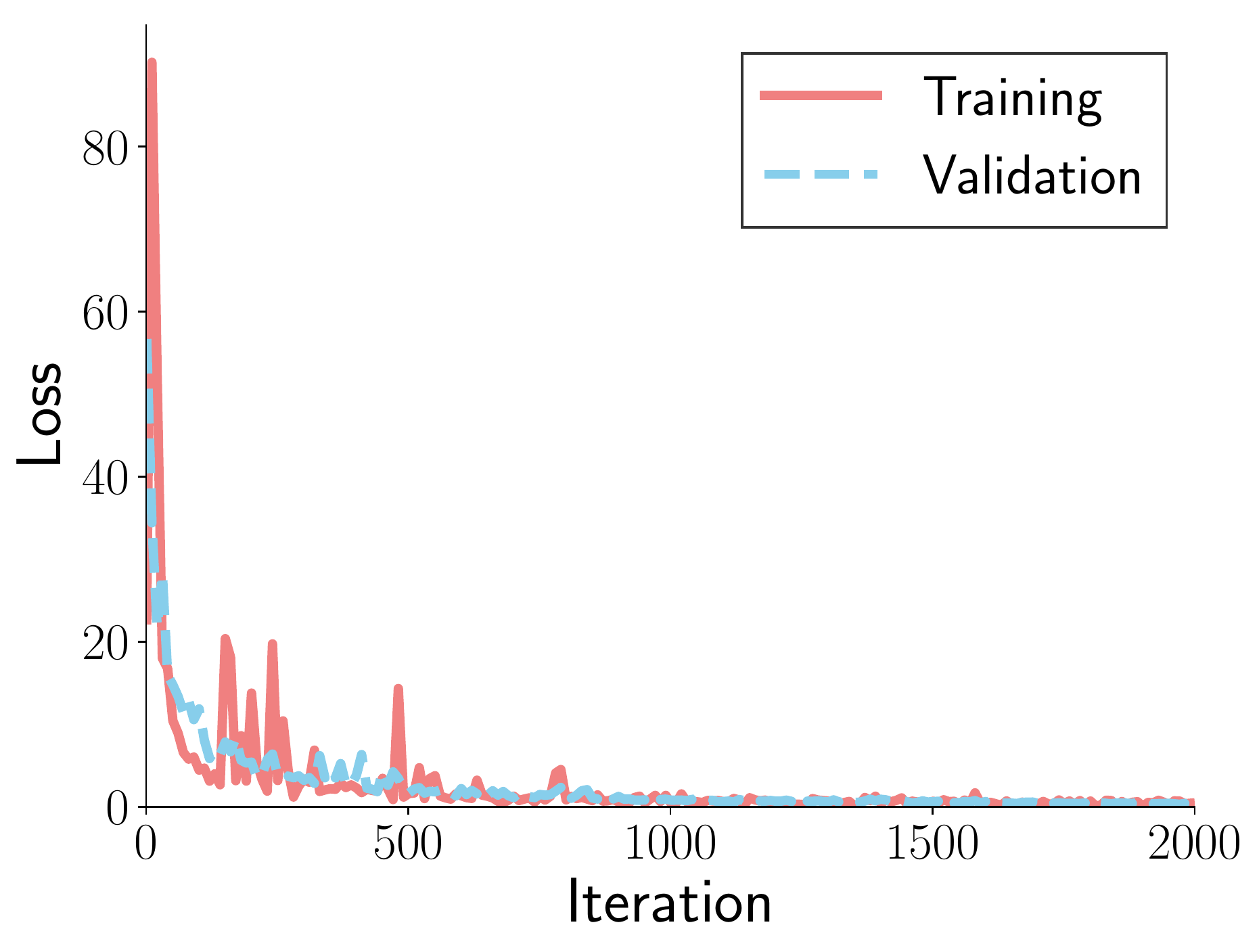}
		\label{fig:loss_set}
	}
	\quad \quad
	\subfigure[Bi-LSTM]{
		\includegraphics[width=0.30\textwidth, keepaspectratio]{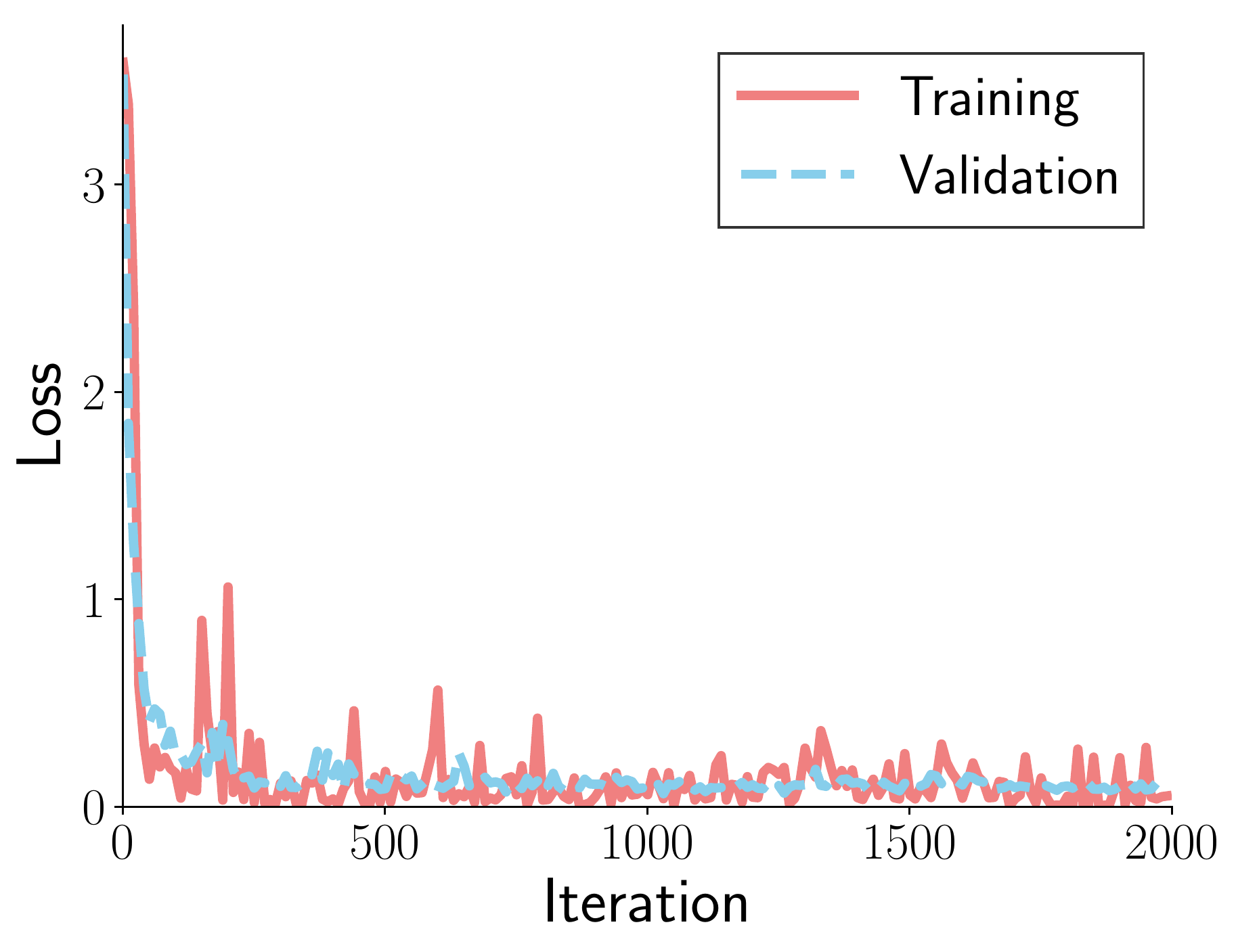}
		\label{fig:loss_lstm}
	}
	\caption{Training and validation losses of ADF and Bi-LSTM for 2,000 iterations.}
	\label{fig:loss}
\end{figure*}

We conducted experiments on learning our models and warm-starting BHO with the following experiment setups. In the end of this section, we will show learned meta-features are helpful to initialize BHO, as shown in \Cref{fig:exp_all}.

\subsection{Experiment Setup}
\label{subsec:setup_exp}

\subsubsection{Collection of Datasets Processing}

We created a collection of datasets for training our model, using eight image datasets: 
\begin{enumerate}[(i)]
	\item Animals with Attributes 2 (AwA2)~\citep{xian2017zero}: 37,322 images of 50 classes,
	\item Caltech-101: 9,146 images of 101 classes,
	\item Caltech-256~\citep{griffin2007caltech}: 30,607 images of 256 classes,
	\item CIFAR-10: 50,000 training images and 10,000 test images of 10 classes,
	\item CIFAR-100: 50,000 training images and 10,000 test images of 100 classes,
	\item Caltech-UCSD Birds-200-2011 (CUB-200-2011)~\citep{WahCUB_200_2011}: 11,788 images of 200 classes,
	\item MNIST: 60,000 training images and 10,000 test images of 10 classes,
	\item PASCAL Visual Objective Classes Challenge 2012 (VOC2012)~\citep{pascal-voc-2012}: 5,717 training images and 5,823 test images of 20 classes. Since VOC2012 is a multi-label dataset, we randomly select one of true labels.
\end{enumerate}

We split each dataset to training, validation and test datasets.
We trained a target model (see the subsequent section) using training datasets, and 
validation datasets are used to measure validation errors for the target model.
Hyperparameter vectors and their validation errors measured by the trained target model are used to train our deep feature and meta-feature extractors.
Moreover, test datasets are considered as new datasets.

If a dataset has a test dataset, we kept it and split training dataset to training and validation datasets.
On the contrary, if a dataset does not include a test dataset, we split the dataset to training, validation, and test datasets.
To magnify the datasets and vary ${\tilde{\theta}_i^{\textrm{c}}}$, 
we subsampled stratified ${\{10\%}$, ${20\%}$, ${\ldots}$, ${90\%}$, ${100\%\}}$ portions of original datasets from the training datasets.
In this paper, we had a collection of 80 datasets, 10 subsampled datasets of 8 image datasets.

\subsubsection{Target Model Setup}

\begin{table*}[t]
	\begin{center}
	\caption{Explanation and range of six-dimensional hyperparameter space. Each space ${\Theta}$ is closed-interval real number space or set. Especially, because a large number of elements in the set cause to generate high-dimensional space internally, \textsf{batch\_size} is on the real number space. To convert a real number to integer, \textsf{batch\_size} is wrapped by integer casting function when the selected hyperparameters are set up to the target model in practice.}
	\label{tab:hyps}
	\begin{tabular}{llll}
		\Xhline{2\arrayrulewidth}\noalign{\smallskip}
		Hyperparameter & Explanation & Range & Type \\
		\noalign{\smallskip}
		\Xhline{1\arrayrulewidth}
		\noalign{\smallskip}
		\textsf{log10\_learning\_rate} & ${\log_{10}}$-scaled initial learning rate & ${[-5.0, 0.0]}$ & Real \\
		\textsf{log10\_decay\_rate} & ${\log_{10}}$-scaled exponential decay & ${[-8.0, -4.0]}$ & Real \\
		& for learning rate & & \\
		\textsf{batch\_size} & batch size & ${[100, 400]}$ & Real \\
		& (wrapped by integer casting) & & (casted to Integer) \\
		\textsf{num\_layers\_conv} & \# of convolutional layers & ${\{1, 2, \ldots, 9\}}$ & Integer \\
		\textsf{num\_layers\_fc} & \# of fully-connected layers & ${\{1, 2, 3\}}$ & Integer \\
		\textsf{dropout\_rate} & dropout rate for dropout layer & ${[0.0, 0.9]}$ & Real \\
		\Xhline{2\arrayrulewidth}
	\end{tabular}
	\end{center}
\end{table*}

\begin{table*}[t]
	\begin{center}
	\caption{Detailed designs of feature extractor and meta-feature extractor. ADF and Bi-LSTM have the following structures. Output channels of convolutional layer, output dimensions of fully-connected layer, and hidden node size of Bi-LSTM are described in parentheses.}
	\label{tab:structure}
	\begin{tabular}{lcc}
		\Xhline{2\arrayrulewidth} \noalign{\smallskip}
		& \textbf{Aggregation of deep features} & \textbf{Bi-directional LSTM} \\
		\noalign{\smallskip}
		\Xhline{1\arrayrulewidth}
		\noalign{\smallskip}
		Deep feature extractor & \multicolumn{2}{c}{3 ${\times}$ convolutional layer with ReLU (32, 64, 32 channels)} \\
		& \multirow{2}{*}{-} & fully-connected layer with ReLU \\
		& & (128 dims) \\
		& \multicolumn{2}{c}{flattening (+ labels)} \\
		Meta-feature extractor & Arithmetic mean aggregation & Bi-LSTM (128 dims) \\
		After extractors & \multicolumn{2}{c}{2 ${\times}$ fully-connected layer with ReLU (256, 256 dims)} \\
		& \multicolumn{2}{c}{fully-connected layer (256 dims)} \\
		\Xhline{2\arrayrulewidth}
	\end{tabular}
	\end{center}
\end{table*}

In this paper we optimized and warm-started convolutional neural network (CNN), created by six-dimensional hyperparameter vector 
${(\textsf{log10\_learning\_rate},}$ ${\textsf{log10\_decay\_rate},}$ ${\textsf{batch\_size},}$ ${\textsf{num\_layers\_conv},}$ ${\textsf{num\_layers\_fc},}$ ${\textsf{dropout\_rate})}$, as shown in \Cref{tab:hyps}.
Assuming that we are given the number of convolutional and fully-connected layers, \textsf{num\_layers\_conv} and \textsf{num\_layers\_fc}, we construct CNN with initial number 32 of ${3 \times 3}$ convolutional filters, max-pooling layers with ${2 \times 2}$ filters, and the fixed node number 256 of fully-connected layers.
The number of convolutional filters are automatically increased and decreased.
For example, if we have five convolutional layers, the number of the filters is ${(32, 64, 128, 64, 32)}$.
All convolutional layers have batch normalization and dropout layer with \textsf{dropout\_rate} is applied in fully-connected layers.
CNN is trained by Adam optimizer with given \textsf{batch\_size} and learning rate.
We scheduled learning rate with exponential decay policy: ${l = 10^{l_0} \exp(-10^{\eta} t)}$,
where ${l_0}$ is \textsf{log10\_learning\_rate}, ${\eta}$ is \textsf{log10\_decay\_rate}, and ${t}$ is the iteration number.

\subsubsection{Feature Extractors Setup}

\begin{algorithm}[t]
	\caption{Bayesian Hyperparameter Optimization with Warm-Starting}
	\label{alg:bho_tip}
	\begin{algorithmic}[1]
		\Require Learned deep feature and meta-feature extractors ${(\calM_{\textrm{df}}, \calM_{\textrm{mf}})}$, target function $\calJ(\cdot)$, limit $T \in \mathbb{N}$, number of initial vectors $k < T$
		\Ensure Best hyperparameter vector $\btheta^{\ast}$
		\State Find ${k}$-nearest neighbors using the learned deep feature and meta-feature extractors, ${(\calM_{\textrm{df}}, \calM_{\textrm{mf}})}$
		\State Obtain ${k}$ historical sets of tuples ${\{ \{ (\btheta_s, \calJ_s^{(1)}) \}_{s=1}^n, \ldots, \{ (\btheta_s, \calJ_s^{(k)}) \}_{s=1}^n\}}$
		\State Initialize an acquired set as an empty set
		\For {$i=1,2, \ldots, k$}
			\State Find the best vector ${\btheta_i^\dagger}$ on grid of the ${i}$-th set of tuples ${\{ (\btheta_s, \calJ_s^{(i)}) \}_{s=1}^n}$
			\State Evaluate $\calJ_i \coloneqq \calJ(\btheta_i^\dagger)$
			\State Accumulate $(\btheta_i^\dagger, \calJ_i)$ into the acquired set
		\EndFor
		\For {$j=k+1,k+2, \ldots, T$}
			\State Estimate a surrogate function $\calM_{\textrm{surrogate}}$ with the acquired set $\{(\btheta_i^\dagger, \calJ_i)\}_{i=1}^{j-1}$
			\State Find $\btheta_j^\dagger = \argmax_{\btheta} a(\btheta | \calM_{\textrm{surrogate}})$.
			\State Evaluate $\calJ_j \coloneqq \calJ(\btheta_j^\dagger)$
			\State Accumulate $(\btheta_j^\dagger, \calJ_j)$ into the acquired set
		\EndFor
		\State \textbf{return} $\btheta^{\ast} = \argmin_{\btheta_j^\dagger \in \{\btheta_1^\dagger, \ldots, \btheta_T^\dagger\}} \calJ_j$
	\end{algorithmic}
\end{algorithm}

To learn a distance function using Siamese network, the network composed of deep feature extractor and meta-feature extractor are used as a wing of Siamese network. As shown in \Cref{tab:structure}, ADF has three convolutional layers with ReLU as deep feature extractor and arithmetic mean aggregation as meta-feature extractor, 
and Bi-LSTM has three convolutional layers with ReLU, which follows fully-connected layer with ReLU as deep feature extractor and Bi-LSTM as meta-feature extractor. 
Inputs of both ADF and Bi-LSTM have additional dimension for label information as well as the outputs of deep feature extractors, in order to feed useful information which is available to the meta-feature extractor.
And also, they follow two fully-connected layers with ReLU and one fully-connected layer. 
As a result, the last output of the last fully-connected layer is used as meta-feature vector.

\subsubsection{Bayesian Optimization Setup}

We employed Bayesian optimization package, \texttt{GPyOpt}~\citep{gpyopt} in BHO.
GP regression with automatic relevance determination (ARD) Mat\'{e}rn 5/2 kernel is used as surrogate function, and EI and GP-UCB are used as acquisition functions.
All experiments are repeated five times.

\subsection{Siamese Network Training}
\label{subsec:training_siamese}

\begin{figure*}[t]
	\centering
	\subfigure[AwA2]{
		\includegraphics[width=0.30\textwidth, keepaspectratio]{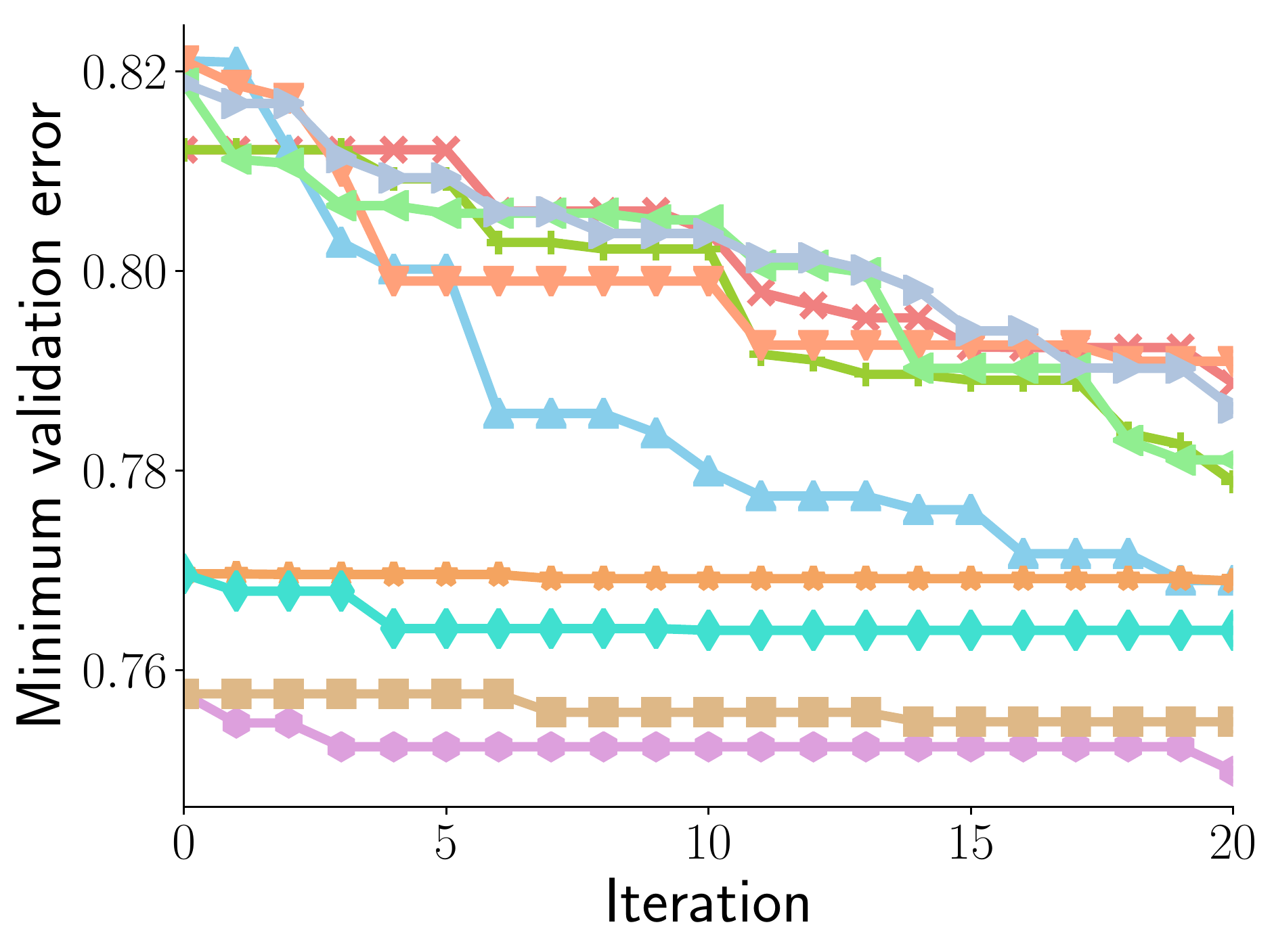}
		\label{fig:bho_awa2}
	}
	\subfigure[Caltech-101]{
		\includegraphics[width=0.30\textwidth, keepaspectratio]{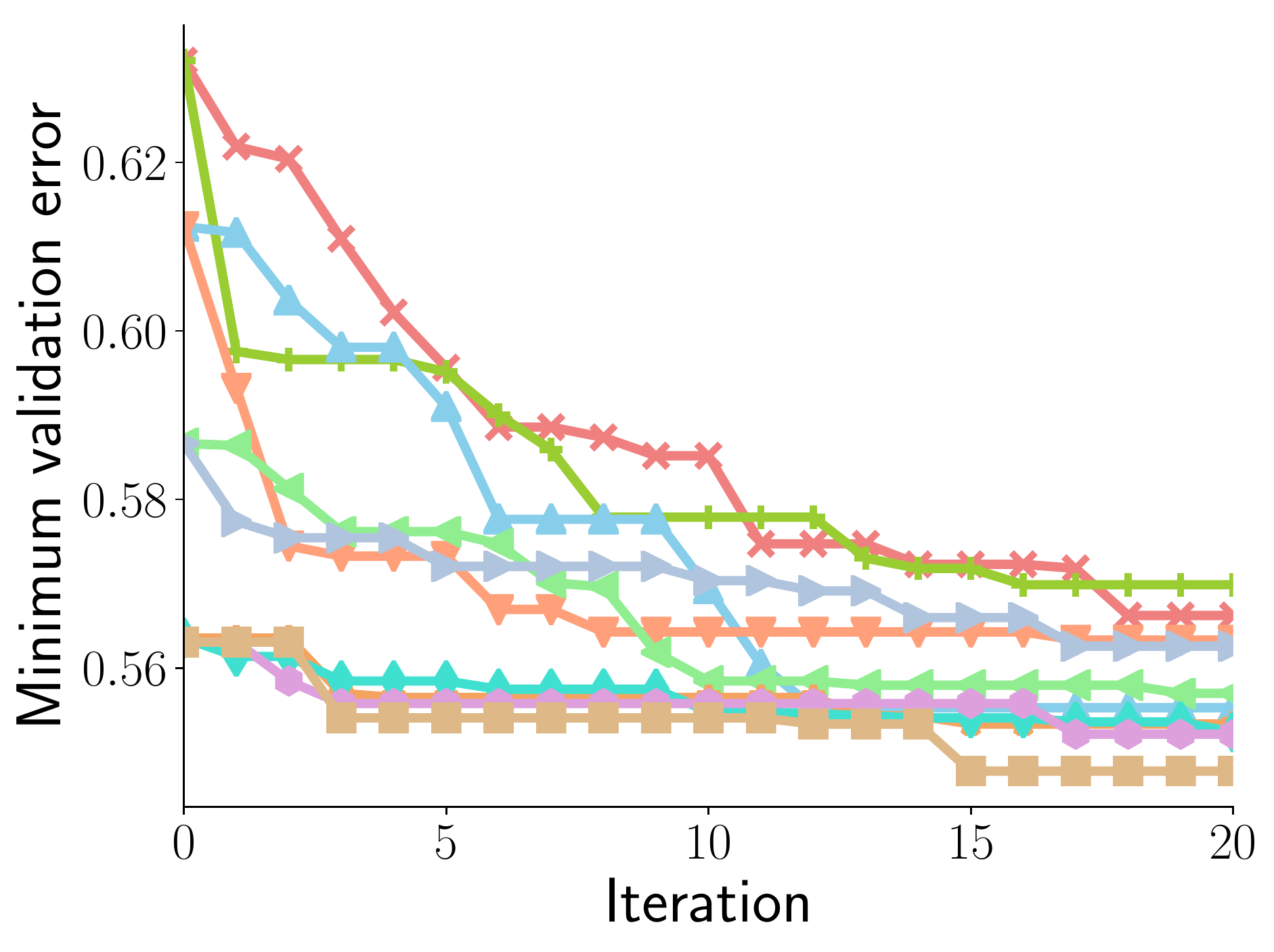}
		\label{fig:bho_caltech101}
	}
	\subfigure[Caltech-256]{
		\includegraphics[width=0.30\textwidth, keepaspectratio]{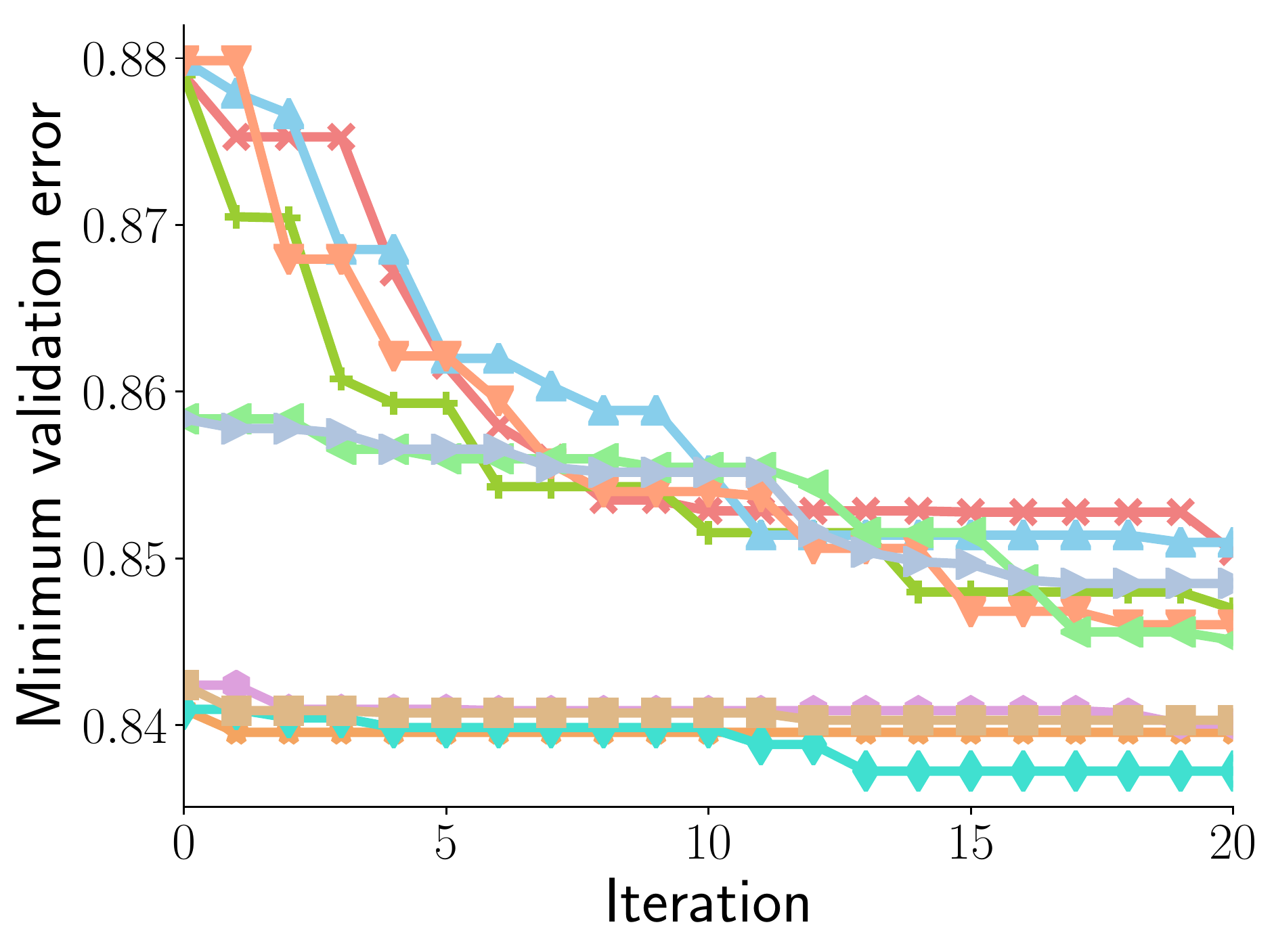}
		\label{fig:bho_caltech256}
	}
	\subfigure[CIFAR-10]{
		\includegraphics[width=0.30\textwidth, keepaspectratio]{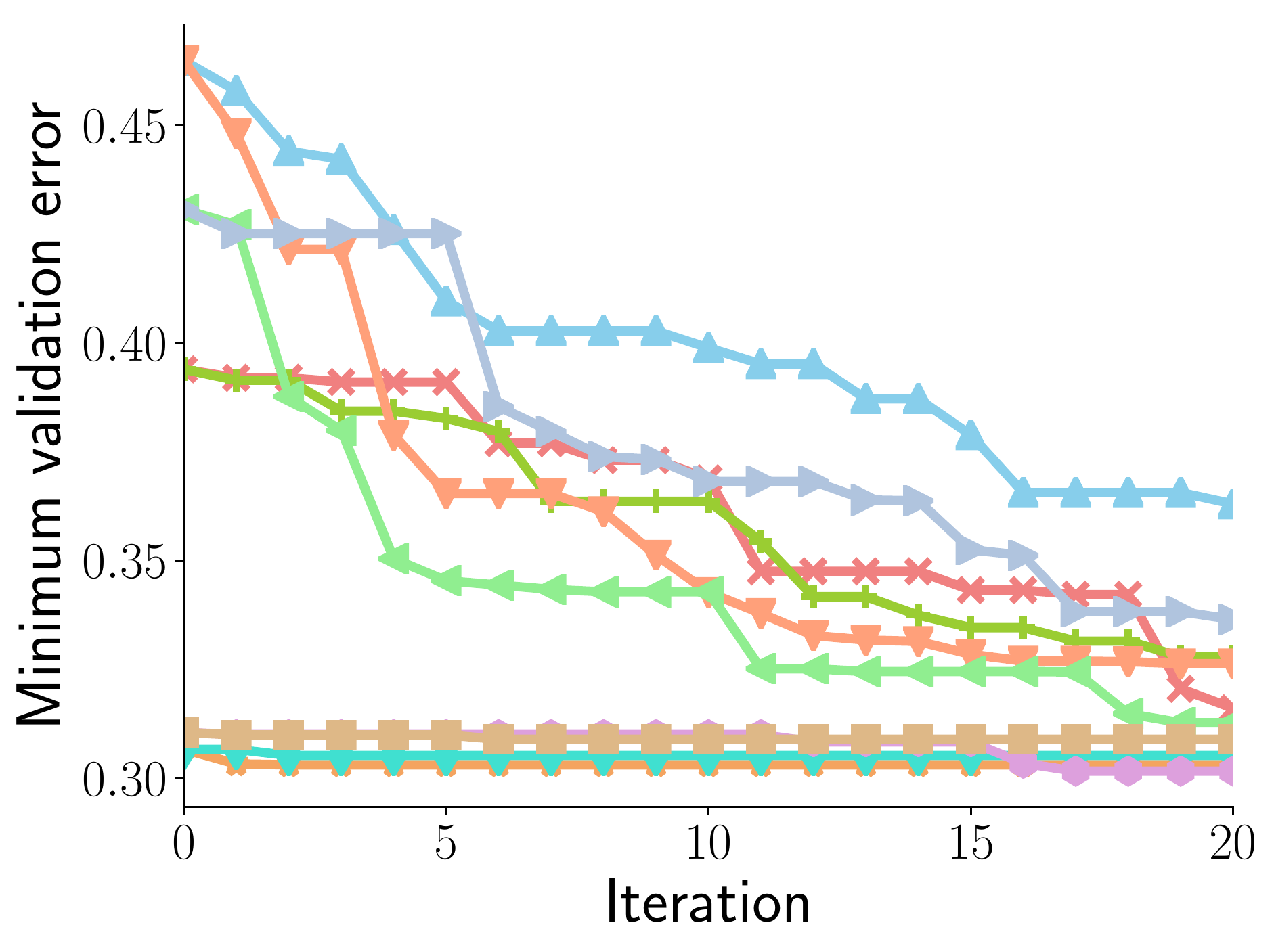}
		\label{fig:bho_cifar10}
	}
	\subfigure[CIFAR-100]{
		\includegraphics[width=0.30\textwidth, keepaspectratio]{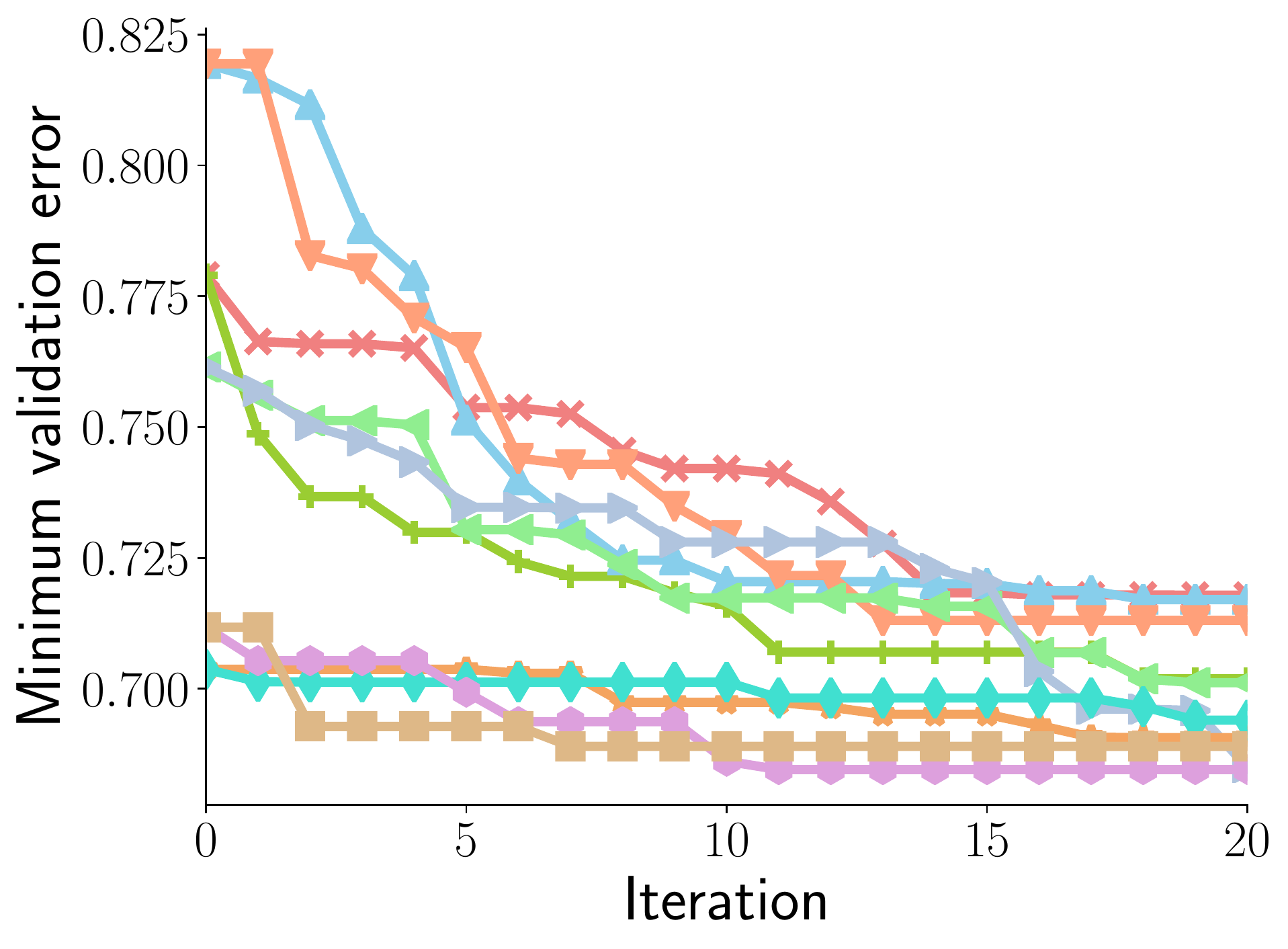}
		\label{fig:bho_cifar100}
	}
	\subfigure[CUB-200-2011]{
		\includegraphics[width=0.30\textwidth, keepaspectratio]{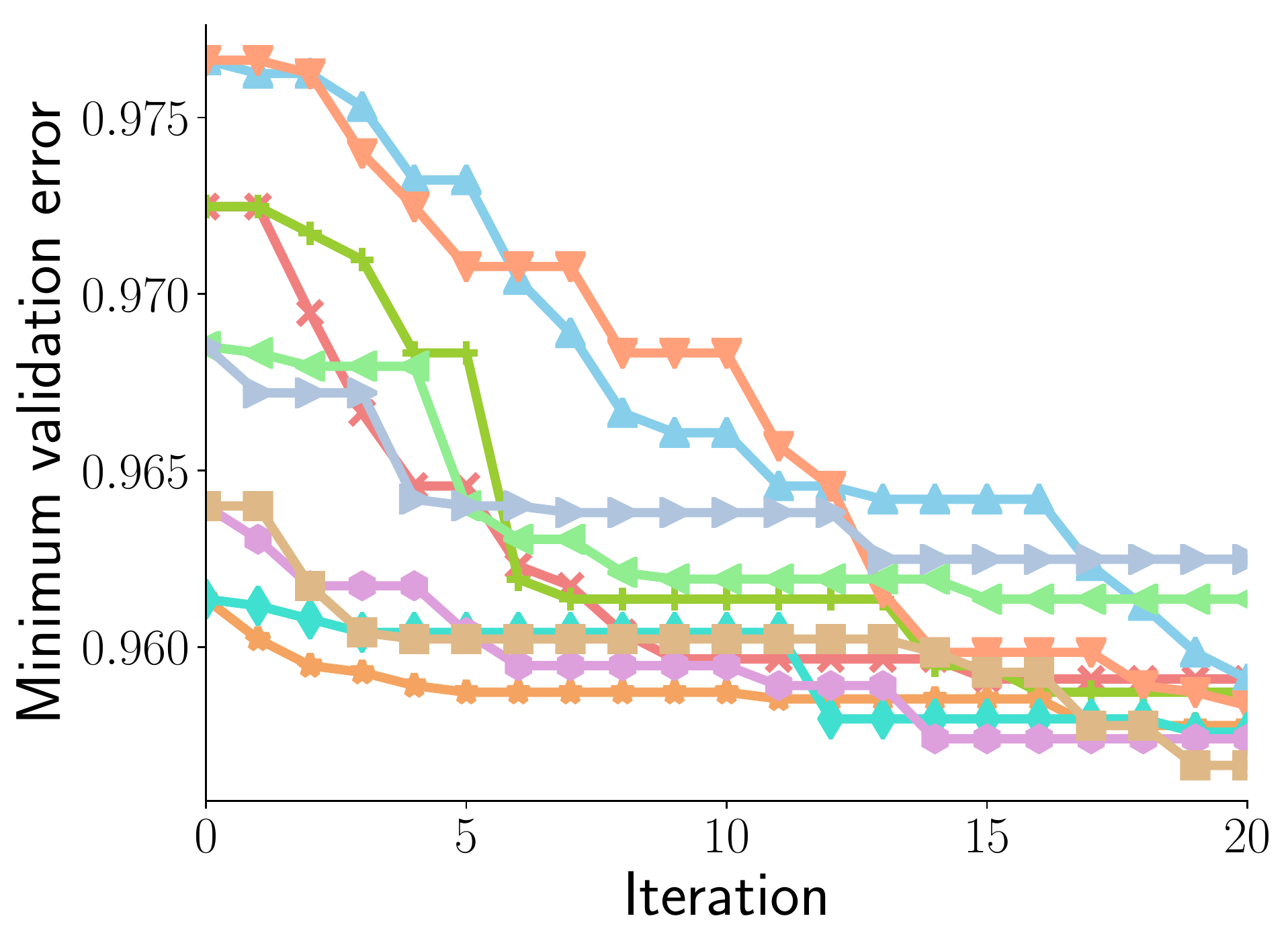}
		\label{fig:bho_cub200_2011}
	}
	\subfigure[MNIST]{
		\includegraphics[width=0.30\textwidth, keepaspectratio]{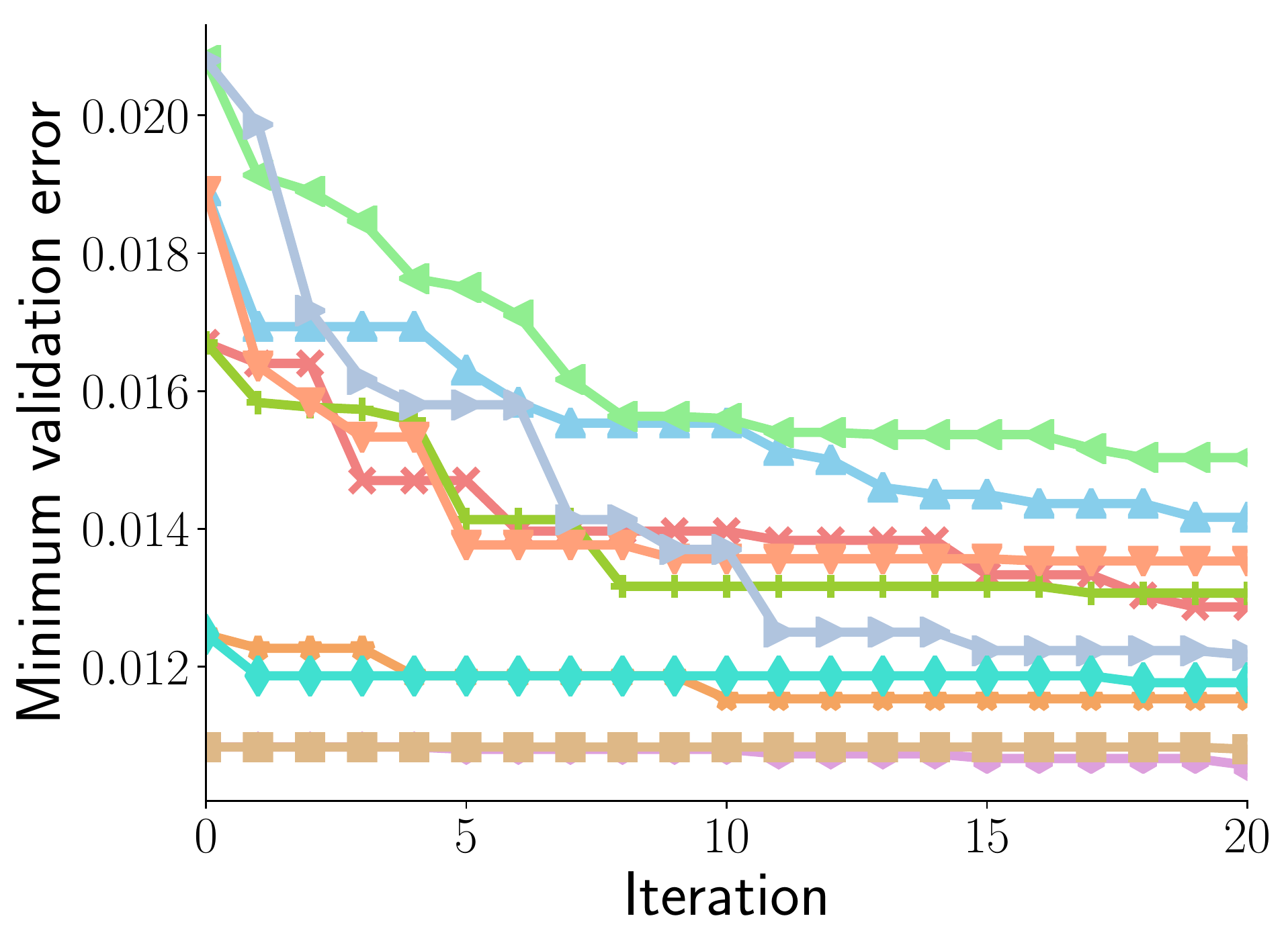}
		\label{fig:bho_mnist}
	}
	\subfigure[VOC2012]{
		\includegraphics[width=0.30\textwidth, keepaspectratio]{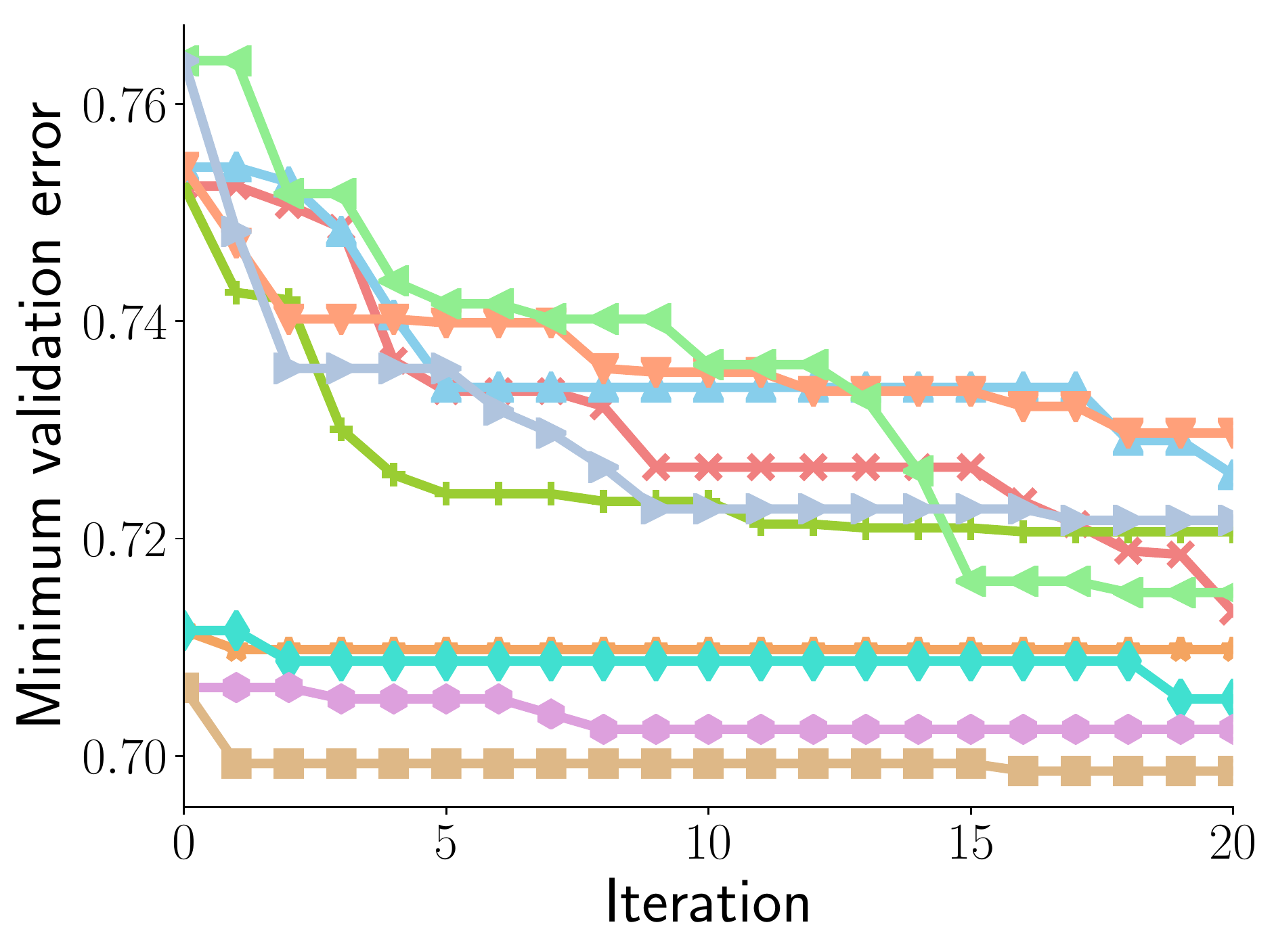}
		\label{fig:bho_voc2012}
	}
	\subfigure{
		\includegraphics[width=0.30\textwidth, keepaspectratio]{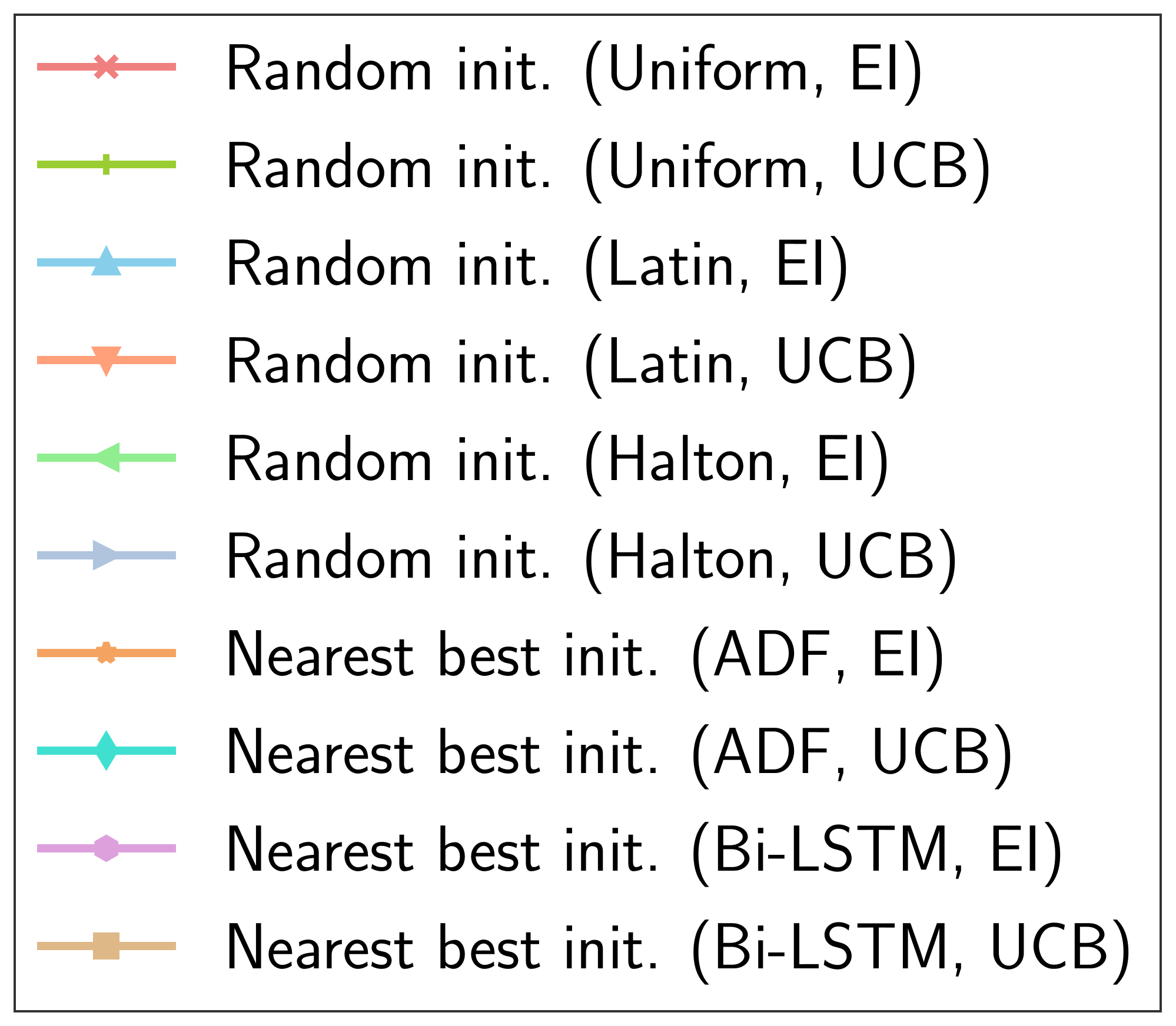}
		\label{fig:bho_legend}
	}
	\caption{BHO for eight test datasets: AwA2, Caltech-101, Caltech-256, CIFAR-10, CIFAR-100, CUB-200-2011, MNIST, and VOC2012. Uniform, Latin, and Halton stand for n\"{a}ive uniform random sampling, Latin hypercube sampling, and quasi-Monte Carlo sampling with Halton sequence, respectively. Standard deviations are not depicted for clarity of figures.}
	\label{fig:exp_all}
\end{figure*}

We trained Siamese networks with ADF and Siamese networks with Bi-LSTM, the structures of which are defined in \Cref{tab:structure}.
They are optimized with Adam optimizer with the number of subsamples 200 and the exponentially decaying learning rate, which has initial learning rate ${10^{-4}}$ and exponential decay rate of learning rate ${10^{-3}}$.

\Cref{fig:loss} shows validation loss as well as training loss are converged to almost zero, and it means our networks can be trained and meta-feature extractor produces an appropriate meta-feature vector that has small loss between target distance and meta-feature distance 
(see the argument for ${\argmin}$ in \Cref{eqn:residual}). 
Learned meta-feature extractor is used in \Cref{subsec:warm_start}.

\subsection{Bayesian Hyperparameter Optimization with Warm-Starting}
\label{subsec:warm_start}

The learned meta-feature extractor which is trained by \Cref{alg:meta_features} is employed in BHO.
BHO with warm-starting, described in \Cref{alg:bho_tip} has inputs: 
(i) learned deep feature and meta-feature extractors ${(\calM_{\textrm{df}}, \calM_{\textrm{mf}})}$,
(ii) target function ${\calJ(\cdot)}$,
(iii) limit ${T}$,
and (iv) the number of initial vectors ${k < T}$.
It follows the similar steps (Lines 9 to 14 of \Cref{alg:bho_tip}) in \Cref{alg:bho}, but initialization steps (Lines 1 to 8 of \Cref{alg:bho_tip}) is different.
Thus, in this section we focus on the initialization step. 

As shown in Line 1 of \Cref{alg:bho_tip}, we find ${k}$-nearest datasets using the meta-feature vectors derived from ${(\calM_{\textrm{df}}, \calM_{\textrm{mf}})}$.
More precisely, we compute the meta-feature vector of new dataset ${\bm_{\textrm{new}}}$ as well as the meta-feature of the datasets in a collection of datasets ${\{ \bm_c \}_{c=1}^\gamma}$ where ${\gamma}$ is the number of datasets in the collection of datasets. 
As mentioned in \Cref{subsec:setup_exp}, ${\gamma = 80}$ in this paper.
Comparing ${\bm_{\textrm{new}}}$ to ${\{\bm_c\}_{c=1}^\gamma}$, we can find ${k}$-nearest datasets.
For ${k}$-nearest datasets, we can select the best hyperparameter vector of dataset ${\calD_i}$ from the historical tuple ${\{(\btheta_s, \calJ_s^{(i)})\}_{s=1}^n}$ for all ${1 \leq i \leq k}$.
Those ${k}$ initial hyperparameter vectors are used to initialize BHO.
In this paper, we used ${3}$ initial hyperparameter vectors for all experiments.

As shown in \Cref{fig:exp_all}, we conducted 10 experiments for each test dataset.
To compare our two methods: 3-nearest best vector initialization predicted by ADF and Bi-LSTM, we used three initialization techniques: (i) na\"{i}ve uniform random sampling (denoted as \emph{uniform}), 
(ii) Latin hypercube sampling (denoted as \emph{Latin}), 
(iii) quasi-Monte Carlo sampling with one of low discrepancy sequences, Halton sequence (denoted as \emph{Halton}).
Moreover, we tested two acquisition functions: EI and GP-UCB (denoted as \emph{UCB}) for five initialization methods.

The compared initialization methods are widely used in initializing BHO. 
In particular, because na\"{i}ve uniform sampling leads to overlapped sampling for each dimension, Latin hypercube sampling and quasi-Monte Carlo sampling are often used.
Two methods sample random vectors that have low overlapped region for each dimension, and they are effective in sampling on high-dimensional space.

All experiments show our methods perform better than other initialization methods.
In addition, Bi-LSTM shows better performance than ADF in most of the experiments.
It implies that Bi-LSTM can learn and extract meta-features well, rather than ADF.

\section{Conclusion}
\label{sec:conclusion}

In this paper we learned meta-feature over datasets using Siamese network with deep feature extractor and meta-feature extractor, in order to warm-start BHO.
We considered identical wings of Siamese network as either ADF or Bi-LSTM, 
and each design shows the network can match pairwise meta-feature distance over datasets with pairwise target distance over them.
Finally, the learned meta-features are used to find a few nearest datasets, and their historical best hyperparameter vectors are utilized in initializing BHO.
Our experiment results for CNNs created by six-dimensional hyperparameter vectors demonstrate that learned meta-features are effective to warm-start BHO.

\bibliographystyle{abbrvnat}
\bibliography{sjc}

\end{document}